%% file: FOREACT3D.tex
\documentclass{article} % For LaTeX2e
\usepackage{iclr2027_conference,times}
\usepackage{booktabs}
\usepackage{multirow}
\usepackage{graphicx}

\input{math_commands.tex}

\usepackage{hyperref}
\usepackage{url}
\usepackage{array}
\usepackage{etoolbox}

\makeatletter
\ifdefined\pdfrunninglinkoff
  \patchcmd{\@outputpage}
    {\@begindvi}
    {\@begindvi\pdfrunninglinkoff}
    {}{\PackageError{ForeAct3D}{Page link patch failed}{Check the output routine.}}
  \patchcmd{\@outputpage}
    {\box\@outputbox}
    {\pdfrunninglinkon\box\@outputbox\pdfrunninglinkoff}
    {}{\PackageError{ForeAct3D}{Body link patch failed}{Check the output routine.}}
\fi
\makeatother

\title{ForeAct3D: Policy-Grounded Future World Modeling for VLA Policies}

\author{
Zhe Tao$^{1}$, Feiran Wang$^{2}$, Gaowen Liu$^{3}$,
Ramana Rao Kompella$^{3}$, Yan Yan$^{2,\dagger}$\\
$^{1}$University of Illinois Urbana-Champaign,
$^{2}$University of Illinois Chicago,
$^{3}$Cisco
}

\iclrfinalcopy % Uncomment for camera-ready version, but NOT for submission.
\newlength{\floatgap}
\AtBeginDocument{%
  \setlength{\floatgap}{1.2\baselineskip}%
  \setlength{\textfloatsep}{\floatgap}%
  \setlength{\floatsep}{\floatgap}%
  \setlength{\intextsep}{\floatgap}%
  \setlength{\dbltextfloatsep}{\floatgap}%
  \setlength{\dblfloatsep}{\floatgap}%
}
\AtBeginEnvironment{table}{%
  \setlength{\abovecaptionskip}{0pt}%
  \setlength{\belowcaptionskip}{6pt}%
}
\AtBeginEnvironment{figure}{%
  \setlength{\abovecaptionskip}{6pt}%
  \setlength{\belowcaptionskip}{0pt}%
}
\begin{document}

\maketitle
\fancyhead{}

\begingroup
\renewcommand{\thefootnote}{\fnsymbol{footnote}}
\footnotetext[2]{Corresponding Author}
% \footnotetext[1]{The code is available at
% \url{https://github.com/anthonytao80-crypto/ForeAct3D}.}
\endgroup

\input{sec/0_abstract}

\input{sec/1_introduction}

\input{sec/2_relatedwork}

\input{sec/3_method}

\input{sec/4_experiment}

\input{sec/5_conclusion}

\clearpage
\input{sec/6_AI_Disclosure}

\input{sec/7_Ethics_Statement}

\input{sec/8_Reproducibility_Statement}

\bibliography{FOREACT3D}
\bibliographystyle{iclr2027_conference}

\input{sec/Appendix}

\end{document}

%% file: math_commands.tex
\usepackage{amsmath,amsfonts,bm}

\def\eqref#1{equation~\ref{#1}}
\def\1{\bm{1}}

\DeclareMathAlphabet{\mathsfit}{\encodingdefault}{\sfdefault}{m}{sl}
\SetMathAlphabet{\mathsfit}{bold}{\encodingdefault}{\sfdefault}{bx}{n}

%% file: sec/0_abstract.tex
\begin{abstract}
Robots need to anticipate how their actions will change the world, since manipulation success hinges on the resulting contacts and object motions. However, existing Vision-Language-Action (VLA) policies that predict future observations from shared features leave the forecast decoupled from the actions the policy will actually execute, and impose no physical constraints on how the scene may evolve. We introduce ForeAct3D, a framework for policy-grounded future world modeling within VLA policies. Learnable geometric queries decode depth, semantic segmentation, and camera pose from the policy representation into current and future semantic 3D scene states, and the future queries are conditioned on the policy-generated action chunk to ground the forecast in the planned interaction. A physical-consistency closure relates the two states through background staticity and instance-level rigidity, and anchors the wrist-camera pose to end-effector kinematics. These objectives shape the shared representation used for action generation during training, and no future prediction is required at inference. Without robot pretraining, ForeAct3D achieves 98.3\% average success on LIBERO and an average task length of 3.73 on CALVIN, outperforming its base policy on every suite. Ablations show that semantic 3D supervision, physical consistency, and action conditioning each improve manipulation performance, and that action conditioning substantially improves future object localization. Real-world experiments on spatial placement, object insertion, and sequential manipulation further raise average success from 6.7\% to 37.8\% over the base policy. The project page and code are available at https://github.com/anthonytao80-crypto/ForeAct3D.
\end{abstract}

%% file: sec/1_introduction.tex
\section{Introduction}
\label{sec:introduction}

\suppressfloats[t]
\input{figs/tex/teaser}

Future world modeling aims to predict how a physical scene will evolve under robot actions. This capability matters for manipulation because the same intended action can succeed or fail depending on contact: a push may slide an object toward its target or tip it over, and an insertion may seat or jam. Anticipating such outcomes requires reasoning about both the current observation and the planned action, since the resulting transitions involve contact, changing occlusions, and object motion~\citep{zhen20243d, wang2026mvista, huang2026pointworld}. Robots therefore need a mechanism that represents the physical world, models its transitions under robot actions, and predicts the resulting future states subject to physical constraints~\citep{lecun2022path, huang2026pointworld}.

Existing approaches to future world modeling for robotic manipulation broadly follow two directions. The first models future observations with action-conditioned video generation~\citep{zhu2025unified,cen2025worldvla,zhu2025irasim}. Such models capture how appearance evolves under actions, but represent the future in image space, leaving the underlying 3D structure and physical constraints implicit. The second extends Vision-Language-Action (VLA) models~\citep{brohan2023rt,kim2024openvla, black2024pi_0} with auxiliary objectives that predict future observations or features from the policy representation~\citep{zhang2026dreamvla, zheng2025flare, sun2026vla, qian2026geopredict}. We adopt the VLA setting because future-state supervision shapes the representation used for action generation while retaining direct action inference without test-time rollouts.

Achieving physically consistent future world modeling within a VLA policy poses three challenges. First, \emph{constructing a task-relevant state} requires recovering 3D geometry from policy features and separating objects, the robot, and the background, so that the predicted state exposes the spatial configuration relevant to manipulation. Second, \emph{constraining state transitions} requires current and future predictions to jointly satisfy physical constraints: viewpoint changes must be reconciled with a static environment, and moving objects must preserve rigid geometry. Predicting each state independently does not enforce these relationships. Third, \emph{grounding the future in action} requires predicting the scene changes induced by the robot's chosen action, since the same scene can evolve differently under different actions. Addressing these challenges calls for a structured state, physical constraints across time, and explicit conditioning on the policy's planned interaction.

To close this gap, we introduce \textbf{ForeAct3D}, an action-conditioned semantic 3D world modeling framework integrated into VLA policies. As illustrated in Figure~\ref{fig:teaser}, ForeAct3D constructs current and future scene states from the representation used for action generation. Learnable geometric queries decode depth, three-class semantic masks, and camera pose, jointly capturing scene geometry and its partition into object, robot, and background regions. During training, a physical-consistency closure relates the two states through background staticity, instance-level rigidity, and end-effector kinematics. Future queries additionally receive the policy-generated action chunk, grounding the predicted transition in the planned interaction. Together, these components train the shared VLA representation to encode a semantic 3D scene and its physically constrained evolution over the action horizon, while retaining direct action inference during deployment.

We evaluate ForeAct3D on LIBERO~\citep{liu2023libero} and CALVIN~\citep{mees2022calvin}, achieving 98.3\% average success and an average task length of 3.73 without robot pretraining. We further assess the predicted states through geometry, segmentation, and localization metrics, and examine the contributions of structured 3D prediction, physical consistency, and action conditioning through ablations. Real-world experiments on spatial placement, insertion, and sequential manipulation tasks confirm these gains on physical hardware. Our contributions are summarized as follows:

\begin{itemize}
    \item We introduce ForeAct3D, a framework for policy-grounded future world modeling within VLA policies that reconstructs current semantic 3D states and predicts future states conditioned on policy-generated actions.

    \item We formulate physical-consistency regularization that relates current and future geometry through background staticity and instance rigidity, while grounding wrist-camera pose in end-effector kinematics.

    \item We demonstrate improved policy performance on LIBERO, CALVIN, and real-world tasks, with component ablations assessing policy gains and prediction ablations showing the benefit of action conditioning for future object localization.
\end{itemize}

%% file: figs/tex/teaser.tex
\begin{figure}[t]
    \centering
    % The supplied PNG is retained verbatim. Typeset panel labels below to
    % avoid generalizing all policy-integrated predictors as action-agnostic.
    \includegraphics[width=\linewidth]{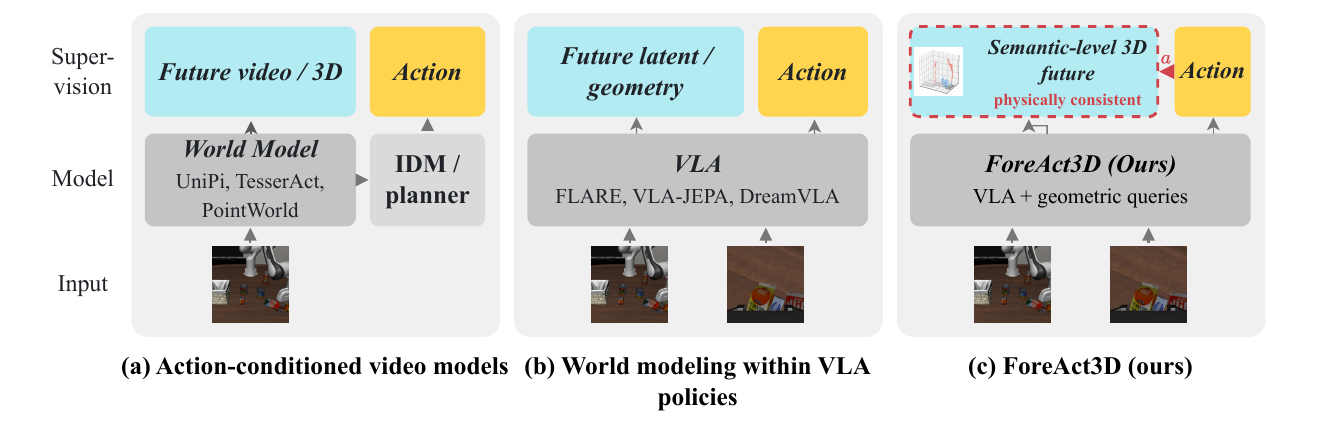}
    \caption{\textbf{Future world modeling for robotic manipulation.}  (a) Predictive models guide control via inverse dynamics or planning.
    (b) VLA policies learn future representations jointly with actions.
    (c) ForeAct3D predicts physically constrained semantic 3D futures conditioned on the policy action.
    Dashed arrows denote supervision linking the predicted scene states to policy learning.}
    \label{fig:teaser}
\end{figure}

%% file: sec/2_relatedwork.tex
\section{Related Work}

Action-conditioned video models predict future observations from the current scene and robot actions~\citep{finn2016unsupervised, finn2017deep, yang2023learning, zhu2025unified, cen2025worldvla}. Approaches include autoregressive prediction over visual and action tokens~\citep{wu2024ivideogpt, wang2025learning} and diffusion-based generation with frame-level action conditioning~\citep{rigter2024avid,zhu2025irasim}. More recently, multi-view prediction and memory mechanisms have improved temporal consistency and enabled policy evaluation within generated rollouts~\citep{guo2026ctrl}. These advances strengthen action controllability and visual coherence, but video prediction alone does not guarantee physically consistent 3D state evolution. For manipulation, visually plausible predictions may remain geometrically inconsistent across views or fail to preserve object structure during interaction. This motivates predicting scene geometry directly and constraining its evolution through physical relationships between current and future states~\citep{wang2026kepler4d}.

\noindent\textbf{Future World Modeling within VLA Policies.}
Another line of work integrates future prediction into VLA policies through auxiliary supervision of features shared with action generation. Prediction targets include latent features~\citep{zheng2025flare, sun2026vla} and visual or geometric observations~\citep{wu2024unleashing,zhen20243d,zhang2026dreamvla,qian2026geopredict,liu2026lift3d}. These objectives encourage anticipation but leave two gaps. First, future prediction alone does not enforce a physically coherent transition from the current scene. Second, shared features alone do not specify which planned action a predicted future corresponds to, motivating explicit action conditioning. ForeAct3D addresses both by jointly predicting current and future semantic 3D states, coupling them through physical-consistency constraints, and conditioning future queries on the policy-generated action chunk. Together, these components supervise structured scene evolution under the planned interaction within the representation used for action generation.

%% file: sec/3_method.tex
\section{Method}
\label{sec:method}

\input{figs/tex/pipeline}

\subsection{Problem Formulation}
\label{sec:problem}

\noindent\textbf{Overview.}
To anticipate the consequences of a planned manipulation, we perform future world modeling through semantic 3D scene prediction within a VLA policy. The model estimates the current scene and predicts its future state over the policy's action horizon. As illustrated in Figure~\ref{fig:pipeline}, structured geometry defines the state, physical-consistency constraints relate the two states across time, and explicit action conditioning grounds the predicted future in the policy's planned action.

At time $t$, the robot receives a language instruction $\ell$ and an observation set $o_t = \{o_t^v\}_{v \in \mathcal{V}}$ containing images from an external agent camera and a robot-mounted wrist camera, where $\mathcal{V}$ denotes these two views. The policy $\pi$ maps $(\ell, o_t)$ to an action chunk $a_{t:t+T}$ of horizon $T$, which also defines the future prediction time $t+T$. This shared horizon aligns future-state supervision with the scene changes induced by the planned action chunk. We predict the scene state at the action horizon's endpoint without generating a sequence of observations for the intermediate action steps.

\noindent\textbf{Semantic 3D scene state.}
To capture the geometry relevant to manipulation, we represent each scene state using depth, three-class semantic masks, and camera pose. Depth encodes spatial structure, semantic masks separate background, robot, and object regions, and camera pose places each view in a common world frame. All objects share a single object class; the prediction neither assigns instance identities nor establishes object correspondence across the current and future states.

For each view $v$, depth $D_t^v$, known intrinsics $K^v$, and camera pose $E_t^v$ define a point cloud $\mathcal{P}_t^v$ by back-projection $\Pi^{-1}$. The semantic mask $S_t^v$ labels these points, so the scene state combines metric geometry with the semantic role of each region:
\begin{equation}
\mathcal{P}_t^v = \Pi^{-1}\!\left(D_t^v, K^v, E_t^v\right),
\qquad
\mathcal{G}_t = \{D_t^v,\, S_t^v,\, E_t^v\}_{v\in\mathcal{V}}.
\end{equation}

ForeAct3D estimates the current state from the instruction and observations, and additionally conditions the future estimate on the planned action chunk. Both estimates live in the same semantic 3D representation, which allows physical constraints to relate the observed scene to its configuration after the action chunk:
\begin{equation}
\hat{\mathcal{G}}_t = f_\theta(o_t, \ell),
\qquad
\hat{\mathcal{G}}_{t+T} = f_\theta(o_t, \ell, a_{t:t+T}).
\end{equation}
The current estimate recovers the scene the policy observes, whereas the future estimate predicts the consequences of the action it generates. This formulation supervises both endpoints of the interaction in a single representation and makes their geometric relationship part of policy learning.

\subsection{Semantic Geometry from Policy Features}
\label{sec:object3d}

\noindent\textbf{Geometric queries.}
To extract a structured scene state from the policy representation, we introduce six learnable queries $q^{(m,\tau)}$, indexed by geometric quantity $m\in\{\mathrm{depth},\mathrm{seg},\mathrm{pose}\}$ and prediction time $\tau\in\{t,t+T\}$. Each query cross-attends to the final backbone features $H_t$ to produce a query feature $z^{(m,\tau)}$; future queries are further conditioned on the action chunk as described in Section~\ref{sec:actionconditioned}. View embeddings are added before decoding: depth and segmentation queries serve both cameras, while pose queries use the wrist-view embedding to predict only the wrist-camera pose. Depth and segmentation share a decoder that attends to intermediate-layer backbone features of the corresponding view, followed by separate output heads. Because all queries read from the policy backbone, geometric supervision directly shapes the representation used for control.

Ground-truth instance IDs provide supervision distinct from the three predicted semantic classes. The rigidity constraint in Section~\ref{sec:closure} uses these IDs to partition the predicted geometry into individual objects and applies Kabsch alignment to each instance using pairs at shared valid pixels. This supplies instance-level physical supervision without requiring the segmentation head to predict object identities or establish object correspondence between the current and future scene states.

\subsection{Physical-Consistency Closure}
\label{sec:closure}

We impose three complementary constraints on predicted geometry and camera pose: background staticity, instance-level rigidity, and end-effector kinematics. The two geometric constraints operate on agent-view point clouds back-projected from predicted depth with ground-truth intrinsics and extrinsics. Since the agent camera is fixed, its extrinsics are identical at both endpoints, and the resulting points are directly comparable in world coordinates. Labeled calibration isolates depth consistency from pose prediction errors, while kinematics supervises the wrist-camera pose.

\noindent\textbf{Background staticity.}
To keep the static environment fixed over time, we compare current and future background points back-projected from predicted depth at the same image pixels. A shared mask retains pixels labeled as valid background at both endpoints, thereby excluding regions occluded by the robot or objects at either time. The background loss $\mathcal{L}_{\mathrm{bg}}$ averages a Smooth-L1 distance $d(\cdot,\cdot)$ over these paired points; the same distance is used in the rigidity term below.

\noindent\textbf{Instance-level rigidity.}
To preserve object shape under motion, we require each instance's current and future geometry to be related by a rigid transform. Ground-truth instance IDs select pixels that belong to the same object $k$ and are valid at both endpoints. For each instance with at least 32 such pixels, we form paired world point sets $\hat{P}_{t,k}$ and $\hat{P}_{t+T,k}$. Kabsch alignment estimates the rigid transform $T_k\in\mathrm{SE}(3)$ between them, and the residual after alignment defines the rigidity loss over the retained instance set $\mathcal{K}$:
\begin{equation}
\mathcal{L}_{\mathrm{rigid}}
=
\frac{1}{|\mathcal{K}|}
\sum_{k\in\mathcal{K}}
w_k\,d\!\left(
T_k\hat{P}_{t,k},
\hat{P}_{t+T,k}
\right).
\end{equation}
Same-pixel pairing approximates temporal correspondence without tracking surface points; it is a short-horizon approximation, since shared instance membership does not guarantee that paired pixels observe the same material point under large motion. The weight $w_k$ down-weights instances likely being grasped: $w_k=0.1$ when the gripper is closed and the current instance centroid lies within 0.10,m of the end effector, and $w_k=1$ otherwise. At the batch level, $\mathcal{L}_{\mathrm{rigid}}$ is averaged with equal weight over samples that have valid future targets and at least one retained instance.

\noindent\textbf{End-effector kinematics.}
To anchor the predicted wrist-camera pose to the robot configuration, we supervise both endpoints using a reference pose from forward kinematics and the fixed hand--eye transform. The pose loss $\mathcal{L}_{\mathrm{ee}}$ sums equally weighted translation MSE and geodesic rotation distance in radians over both prediction times. We convert the predicted 6D rotation into a rotation matrix before computing the loss. Pose supervision is separate from the background and rigidity constraints, which reconstruct world points using predicted depth and labeled camera poses.

\noindent\textbf{Closure.}
The three constraints are combined into a single regularization objective, with coefficients $\lambda_{\mathrm{bg}}$, $\lambda_{\mathrm{rigid}}$, and $\lambda_{\mathrm{ee}}$ controlling their relative contributions:
\begin{equation}
\mathcal{L}_{\mathrm{closure}}
=
\lambda_{\mathrm{bg}}\mathcal{L}_{\mathrm{bg}}
+
\lambda_{\mathrm{rigid}}\mathcal{L}_{\mathrm{rigid}}
+
\lambda_{\mathrm{ee}}\mathcal{L}_{\mathrm{ee}}.
\end{equation}
Together, these constraints encourage a static background, approximately rigid object motion, and wrist-camera poses grounded in robot kinematics. Since geometric predictions are decoded from the policy backbone, these constraints directly regularize its representation for action generation.

\input{tab/libero}

\subsection{Action-Conditioned State Prediction}
\label{sec:actionconditioned}

The current scene alone does not determine an interaction's outcome: different actions can produce different future states from the same observation. The future branch therefore receives the policy-generated action chunk, explicitly tying its prediction to the planned interaction.

\noindent\textbf{Action-conditioned prediction.}
To inject the planned action into future prediction, we project the action chunk into the query feature space with a small network $\phi$. After all six queries have read scene information from the backbone (Section~\ref{sec:object3d}), only the three future queries are additionally conditioned on the action. Each future query cross-attends to the concatenation of the projected backbone features $\bar H_t$ and the action features $\phi(a_{t:t+T})$:
\begin{equation}
\tilde{z}^{(m,t+T)}
=
\operatorname{CrossAttn}\!\left(
z^{(m,t+T)};\,\bar H_t,\,\phi(a_{t:t+T})
\right).
\end{equation}
This step fuses scene context with the planned action into the future query features $\tilde z^{(m,t+T)}$. Current queries skip this step and keep their scene-only features $z^{(m,t)}$. View embeddings are then added to both sets of queries before decoding depth, segmentation, and wrist-camera pose, as in Section~\ref{sec:object3d}.

%% file: figs/tex/pipeline.tex
\begin{figure}[t]
    \centering
    \includegraphics[width=\linewidth, trim=0 0 0 2pt, clip]{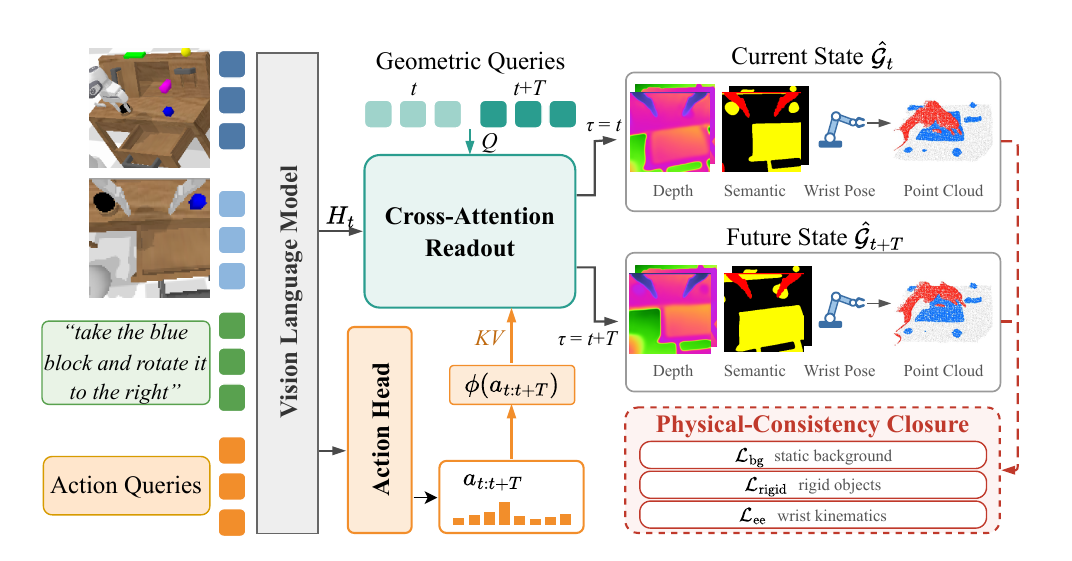}
    \caption{
        \textbf{Overview of the proposed framework.}
        ForeAct3D integrates semantic 3D prediction, physical-consistency closure, and action-conditioned future forecasting into a unified VLA policy for predicting task-relevant scene evolution under the planned action chunk.}
    \label{fig:pipeline}
    
\end{figure}

%% file: tab/libero.tex
\begin{table}[!t]
\centering
\caption{Manipulation success rates (\%) on LIBERO.
Results cover four task suites and their average.
Methods are grouped by training protocol and sorted within groups by average success.}
\label{tab:libero}
\small
\setlength{\tabcolsep}{2pt}
\begin{tabular*}{\linewidth}{
@{\extracolsep{\fill}}
>{\raggedright\arraybackslash}p{0.36\linewidth}
lccccc@{}
}
\toprule
Method & Venue, Year & Spatial & Object & Goal & Long & Avg. \\
\midrule
\multicolumn{7}{l}{\textit{One model per suite}} \\

DreamVLA~\citep{zhang2026dreamvla}
& NeurIPS'25
& 97.5 & 94.0 & 89.5 & 89.5 & 92.6 \\

QDepth-VLA~\citep{li2025qdepth}
& arXiv'26
& 97.6 & 96.6 & 95.2 & 90.0 & 94.9 \\

UniVLA~\citep{bu2025univla}
& RSS'25
& 96.5 & 96.8 & \underline{95.6} & 92.0 & 95.2 \\

ST4VLA~\citep{ye2026st4vla}
& ICLR'26
& \underline{98.0} & \underline{99.0}
& 93.8 & \underline{92.6} & \underline{95.9} \\

\textbf{Spatial Forcing}~\citep{li2026spatial}
& ICLR'26
& \textbf{99.4} & \textbf{99.6}
& \textbf{98.8} & \textbf{96.0} & \textbf{98.5} \\

\midrule
\multicolumn{7}{l}{\textit{One model for all suites}} \\

UD-VLA~\citep{chen2026unified}
& ICLR'26
& 94.1 & 95.7 & 91.2 & 89.6 & 92.7 \\

Evo-1~\citep{lin2026evo}
& CVPR'26
& 92.7 & 97.7 & 96.3 & 92.3 & 94.8 \\

UniVLA~\citep{wang2026unified}
& ICLR'26
& 95.4 & 98.8 & 93.6 & 94.0 & 95.5 \\

StarVLA-OFT~\citep{community2026starvla}
& arXiv'26
& 97.8 & 98.6 & 96.2 & 93.8 & 96.6 \\

FASTer~\citep{liu2025faster}
& ICLR'26
& \underline{98.0} & \textbf{99.4}
& \textbf{98.6} & \underline{95.4} & \underline{97.9} \\

\textbf{Ours}
& --
& \textbf{99.2} & \underline{99.2}
& \underline{98.4} & \textbf{96.2} & \textbf{98.3} \\

\bottomrule
\end{tabular*}
\end{table}

%% file: sec/4_experiment.tex
\section{Experiments}
\label{sec:experiments}

We report manipulation results on LIBERO and CALVIN (Section~\ref{sec:results}) and assess predicted geometry, semantics, and future object localization (Section~\ref{sec:worldmodel}). We ablate individual components (Section~\ref{sec:ablation}) and evaluate three real-world tabletop manipulation tasks (Section~\ref{sec:realworld}).

\subsection{Manipulation Performance}
\label{sec:results}

\noindent\textbf{LIBERO.} Table~\ref{tab:libero} reports success rates across four suites under two training protocols: one model per suite or one model for all suites. Using a single model for all suites, ForeAct3D achieves 98.3\% average success, the highest among listed methods under this protocol. It outperforms its base policy StarVLA-OFT by 1.7 percentage points, with gains across all four suites and the largest on Goal and Long, supporting the benefit of structured 3D prediction without suite-specific adaptation.

\noindent\textbf{CALVIN.}
We evaluate long-horizon language-conditioned manipulation in the ABC$\rightarrow$D setting, focusing on methods without robot pretraining. ForeAct3D and StarVLA-OFT share the same training configuration, with the proposed geometric modeling components as the primary difference. ForeAct3D increases average task length from 2.94 to 3.73 (Table~\ref{tab:calvin}), surpassing all listed methods without robot pretraining, including Seer trained from scratch (3.64).

\input{tab/calvin}

\subsection{Geometry, Semantics, and Future Localization}
\label{sec:worldmodel}

We evaluate three aspects of the predicted state separately: geometry back-projected from predicted depth within ground-truth (GT) object regions, three-class semantic segmentation, and future object localization from predicted depth and semantic masks. Table~\ref{tab:worldmodel} summarizes the results; full metrics and protocols appear in Appendices~\ref{app:evaluation} and~\ref{app:semantic_results}. In both localization settings, GT instance IDs define which objects are evaluated; the predicted-mask setting therefore tests localization given predicted semantic support, not autonomous instance discovery.

\input{tab/future_3d_world_model}

\noindent\textbf{Geometry within known object regions.}
Table~\ref{tab:worldmodel}(a) isolates depth-derived geometry from segmentation quality. On CALVIN, ForeAct3D achieves lower future instance-level error (Inst-CD 0.0137 vs. 0.0139\,m) and higher Recall@2cm (92.8\% vs. 75.8\%) than CLOVER, whereas CLOVER has slightly lower scene-level error (Scene CD 0.0195 vs. 0.0206\,m). The prediction horizons differ ($T=8$ for ForeAct3D, $5$ for CLOVER, and $3$ for DreamVLA), so the comparison reflects each method's accuracy at its own horizon rather than a matched-horizon advantage; note that ForeAct3D predicts the farthest ahead. Table~\ref{tab:geometry_full} further shows that ForeAct3D's GT-mask Recall@2cm decreases only modestly from 94.2\% (current) to 92.8\% (future). These scores are conditional: instances with insufficient valid depth are excluded without penalty, and predicted segmentation is not evaluated.

\noindent\textbf{Semantic regions across time.}
Table~\ref{tab:worldmodel}(b) evaluates the predicted segmentation directly. From the current to the future state, mean IoU decreases only slightly, from 89.3\% to 88.8\% on LIBERO and from 88.0\% to 85.8\% on CALVIN. Per-class scores in Table~\ref{tab:semantic_quality} are less uniform, however: on LIBERO Spatial, future object IoU is only 45.7\% while background and robot IoU reach 91.5\% and 95.7\%. These averages therefore do not guarantee accurate object masks on every task.

\noindent\textbf{Localization with predicted semantic support.}
On CALVIN, replacing GT masks with predicted masks increases translation error from 1.25 to 3.49\,cm and reduces Recall@2cm from 92.8\% to 65.9\%, exposing segmentation errors that the GT-mask evaluation hides. On LIBERO, predicted masks yield 0.87\,cm and 94.9\%, versus 1.06\,cm and 94.0\% with GT masks. This lower conditional error does not indicate better scene recovery: predicted masks change which points enter the centroid computation, and may drop hard pixels that GT masks retain.

\subsection{Real-World Manipulation}
\label{sec:realworld}

We evaluate ForeAct3D on three tabletop manipulation tasks using the AgileX platform (Figure~\ref{fig:visualization_real}). We collect 50 demonstrations per task and evaluate each configuration over 15 trials per task. Depth camera measurements and SAM3~\citep{carion2026sam} segmentation provide geometric supervision.

\input{tab/realworld}

\noindent\textbf{Task design.}
The tasks probe spatial relations, precise placement, and sequential dependencies. \emph{Moving the mug to the yellow area} tests target-region retention: success requires the mug to remain inside the area at termination, whereas reach records whether it entered at any point. This distinguishes transiently reaching a target from maintaining the intended configuration after the interaction ends. \emph{Placing the pen in the mug} requires accurate relative geometry of the pen, mug opening, and gripper. The long task, \emph{placing the mug in a container and then the pen in the mug}, tests sequential dependence: the first stage determines the target pose for the second. Full success requires both stages; mug placement is also reported as intermediate progress within the same trials.

\noindent\textbf{Result analysis.}
Table~\ref{tab:realworld} shows average success increasing from 6.7\% to 22.2\%, 26.7\%, and 37.8\% as components are added, corresponding to 3/45 successes for StarVLA-OFT and 17/45 for ForeAct3D. Semantic 3D prediction provides the largest gain, particularly for placement through elevated openings, where depth informs placement position and release height. Physical-consistency closure further improves pen placement from 7/15 to 9/15 and first-stage mug placement from 5/15 to 6/15, consistent with benefits from more coherent geometry. With action conditioning, all 5/5 trials reaching the yellow area retain the mug there, compared with 2/6 without it. Long-task completion rises from 1/15 to 3/15 while first-stage mug placement remains at 6/15. The gain occurs in the dependent second stage, whose target mug pose is established by preceding actions, supporting action-conditioned prediction when later actions depend on earlier outcomes.

\subsection{What Makes the Predictive State Useful?}
\label{sec:ablation}
\label{sec:action_prediction}

All ablation variants share the same backbone, training data, schedule, and evaluation protocol. In the action-conditioning ablation, the geometric queries and physical-consistency closure are kept unchanged and only the action input to the future queries is removed, isolating its contribution.

\input{figs/tex/visualization}

\input{tab/ablation}

\noindent\textbf{From geometric supervision to the full model.}
Table~\ref{tab:ablation}(a) shows a progressive improvement from StarVLA-OFT (96.6\%) to semantic 3D prediction (97.3\%), physical-consistency closure (97.9\%), and action conditioning (98.3\%). Each addition yields a further gain, indicating that the three components contribute complementarily to manipulation performance.

\noindent\textbf{Contributions of physical constraints.}
Table~\ref{tab:ablation}(a) ablates individual closure terms on CALVIN, retaining semantic 3D prediction and action conditioning. Removing background staticity, instance-level rigidity, or kinematic supervision reduces average task length from 3.73 to 3.65, 3.55, and 3.68, respectively; removing all three reduces it to 3.45. Removing rigidity causes the largest individual drop, supporting the importance of object shape preservation for sequential manipulation.

\noindent\textbf{Predicted states during execution.}
Figure~\ref{fig:visualization} shows current observations, predicted future semantic 3D states, and the resulting executions for drawer opening in LIBERO-Goal and light switching in CALVIN; the predicted object and robot geometry tracks the actual scene evolution. Note that future predictions serve only as training supervision and are not used to select or refine actions at inference.

\noindent\textbf{Action input and object motion.}
Table~\ref{tab:ablation}(b) compares future geometry with and without action conditioning. On CALVIN, removing the action input raises Inst-CD from 0.0137 to 0.0183\,m and translation error from 1.25 to 1.77\,cm, while Recall@2cm falls from 92.8\% to 79.9\%. On LIBERO, suite-averaged translation error rises from 1.06 to 2.59\,cm and recall falls from 94.0\% to 88.4\%, with consistent improvements across all four suites (Appendix~\ref{app:action_ablation}). These results support conditioning future queries on planned actions to anticipate object motion.

%% file: tab/calvin.tex
\begin{table}[t]
\centering
\caption{Long-horizon performance on CALVIN ABC$\rightarrow$D benchmark.
Methods are grouped by training protocol and sorted by average rollout length.
$^{*}$ marks results reproduced by us.}
\label{tab:calvin}
\small
\setlength{\tabcolsep}{2pt}
\begin{tabular*}{\linewidth}{@{\extracolsep{\fill}}llcccccc@{}}
\toprule
Method & Venue, Year & Avg. Len.
& Task 1 & Task 2 & Task 3 & Task 4 & Task 5 \\
\midrule
\multicolumn{8}{l}{\textit{With robot pretraining}} \\

Retrieval-OpenVLA~\citep{zhang2026retrieval}
& CVPR'26
& 3.53 & 92.7 & 79.6 & 70.5 & 58.3 & 51.9 \\

Dita~\citep{hou2025dita}
& ICCV'25
& 3.61 & 94.5 & 82.5 & 72.8 & 61.3 & 50.0 \\

GHIL-Glue~\citep{hatch2025ghil}
& ICRA'25
& 3.69 & 95.2 & 88.5 & 72.3 & 62.9 & 49.8 \\

ProgressVLA~\citep{yan2026progressvla}
& arXiv'26
& 3.73 & 95.2 & 84.8 & 73.6 & 67.2 & 52.0 \\

MoDE~\citep{reuss2025efficient}
& ICLR'25
& 4.01 & 96.2 & 88.9 & 81.1 & 71.8 & 63.5 \\

Seer-Large~\citep{tian2025predictive}
& ICLR'25
& \underline{4.28}
& \underline{96.3} & \underline{91.6}
& \underline{86.1} & \underline{80.3} & \underline{74.0} \\

\textbf{DreamVLA}~\citep{zhang2026dreamvla}
& NeurIPS'25
& \textbf{4.44}
& \textbf{98.2} & \textbf{94.6}
& \textbf{89.5} & \textbf{83.4} & \textbf{78.1} \\

\midrule
\multicolumn{8}{l}{\textit{Without robot pretraining}} \\

Dita
& ICCV'25
& 2.38 & 89.5 & 63.3 & 39.8 & 27.3 & 18.5 \\

StarVLA-OFT$^{*}$
& arXiv'26
& 2.94 & 86.4 & 70.2 & 55.5 & 45.2 & 36.2 \\

ProgressVLA
& arXiv'26
& 3.24 & 89.4 & 76.8 & 63.0 & 52.2 & 43.1 \\

MoDE
& ICLR'25
& 3.39 & 91.5 & 79.2 & 67.3 & 55.8 & 45.3 \\

CLOVER~\citep{bu2024closed}
& NeurIPS'24
& 3.53
& \textbf{96.0} & \underline{83.5}
& 70.8 & 57.5 & 45.4 \\

Seer (scratch)
& ICLR'25
& \underline{3.64}
& 93.0 & 82.4
& \underline{72.3} & \underline{62.6} & \underline{53.3} \\

\textbf{Ours}
& --
& \textbf{3.73}
& \underline{94.2} & \textbf{85.3}
& \textbf{74.4} & \textbf{64.4} & \textbf{54.8} \\

\bottomrule
\end{tabular*}
\end{table}

%% file: tab/future_3d_world_model.tex
\begin{table}[t]
\centering
\small
\caption{\textbf{Geometry, semantics, and future localization.} (a) CALVIN geometry; instance metrics use GT masks and horizons differ by method. (b) ForeAct3D semantic mIoU and localization at $t+T$. GT / Pred denotes GT / predicted semantic masks; both retain GT instance IDs. Recall is in percent. Details and protocol qualifications are in Appendix~\ref{app:semantic_results}.}
\label{tab:worldmodel}
\textbf{(a) Future geometry: scene and GT-mask object metrics}\par\smallskip
\begin{tabular}{@{}lccccc@{}}
\toprule
Method & Horizon & Scene CD (m) $\downarrow$ & Inst-CD (m) $\downarrow$ & Trans (cm) $\downarrow$ & Recall $\uparrow$ \\
\midrule
CLOVER & 5 & \textbf{0.0195} & 0.0139 & 1.741 & 75.82 \\
DreamVLA & 3 & 0.1350 & 0.1455 & 20.61 & 10.73 \\
ForeAct3D & 8 & 0.0206 & \textbf{0.01371} & \textbf{1.245} & \textbf{92.78} \\
\bottomrule
\end{tabular}
\par\medskip
\textbf{(b) Semantic prediction and localization with predicted masks}\par\smallskip
\begin{tabular}{@{}lccc@{}}
\toprule
Benchmark & mIoU (\%), $t$ / $t+T$ $\uparrow$ & Trans (cm), GT / Pred $\downarrow$ & Recall, GT / Pred $\uparrow$ \\
\midrule
LIBERO & 89.3 / 88.8 & 1.06 / 0.87 & 94.0 / 94.9 \\
CALVIN & 88.0 / 85.8 & 1.25 / 3.49 & 92.8 / 65.9 \\
\bottomrule
\end{tabular}
\end{table}

%% file: tab/realworld.tex
\begin{table}[t]
\centering
\caption{\textbf{Real-world cumulative ablations.} Rates (\%) are followed by counts in parentheses, with 15 trials per configuration and task. Reach means the object entered the yellow area at least once; success requires it to remain there at termination. First stage records long-task mug placement.}
\label{tab:realworld}
\small
\begin{tabular*}{\linewidth}{@{\extracolsep{\fill}}lccccc@{}}
\toprule
& \multicolumn{2}{c}{Yellow area} & Pen in mug & \multicolumn{2}{c}{Long task} \\
\cmidrule(lr){2-3}\cmidrule(lr){4-4}\cmidrule(lr){5-6}
Configuration & Success & Reach & Success & Full & First stage \\
\midrule
StarVLA-OFT & 0.0 (0/15) & 6.7 (1/15) & 20.0 (3/15) & 0.0 (0/15) & 13.3 (2/15) \\
+ Semantic 3D & 13.3 (2/15) & 33.3 (5/15) & 46.7 (7/15) & 6.7 (1/15) & 33.3 (5/15) \\
+ Physical consistency & 13.3 (2/15) & \textbf{40.0 (6/15)} & \textbf{60.0 (9/15)} & 6.7 (1/15) & \textbf{40.0 (6/15)} \\
+ Action conditioning (Ours) & \textbf{33.3 (5/15)} & 33.3 (5/15) & \textbf{60.0 (9/15)} & \textbf{20.0 (3/15)} & \textbf{40.0 (6/15)} \\
\bottomrule
\end{tabular*}
\end{table}

%% file: figs/tex/visualization.tex
\begin{figure}[t]
    \centering
    \includegraphics[width=\linewidth]{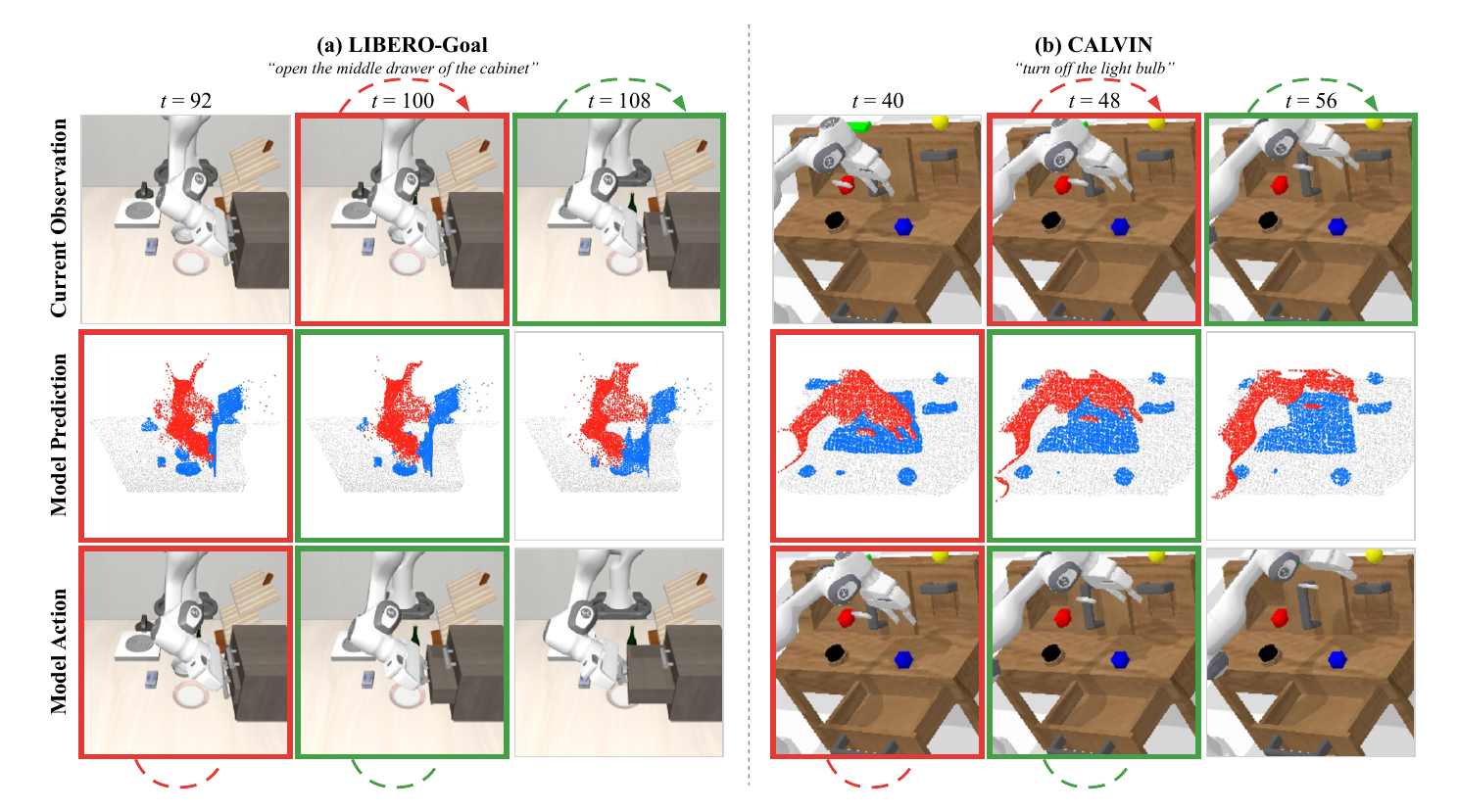}
    \caption{\textbf{Scene prediction and policy execution in simulation.} Examples show opening a drawer in LIBERO-Goal and turning off a light in CALVIN. Rows present current observations, predicted semantic 3D states, and execution results. Matching colored borders link corresponding observation and execution frames across successive steps of each manipulation sequence.}
    \label{fig:visualization}
    \par\vspace{1.2\baselineskip}
    \input{figs/tex/visualization_real}
\end{figure}

%% file: figs/tex/visualization_real.tex
    \centering
    \includegraphics[width=\linewidth]{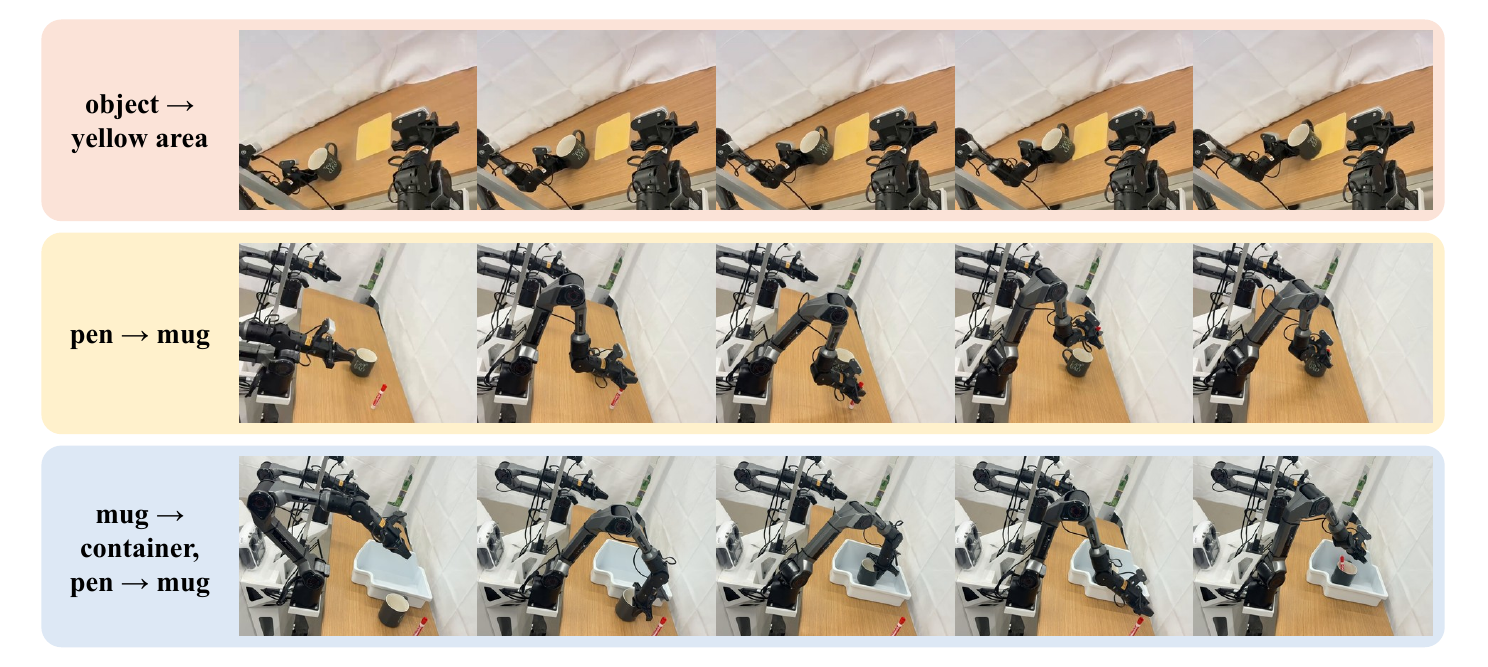}
    \caption{\textbf{Real-world task execution.} Rows show moving an object to the yellow area, placing a pen in a mug, and placing the mug in a container before inserting the pen. Frames proceed from left to right and illustrate the spatial goals and sequential structure of the three tasks.}
    \label{fig:visualization_real}

%% file: tab/ablation.tex
\begin{table}[t]
\centering
\caption{\textbf{Policy and prediction ablations.}
(a) Cumulative policy ablations on LIBERO and closure ablations on CALVIN.
CALVIN variants retain semantic 3D prediction and action conditioning.
Scores report LIBERO success (\%, averaged over four suites) and mean CALVIN task length.
(b) Future geometry with and without action conditioning.
Detailed metrics appear in Appendix~\ref{app:action_ablation}.}
\label{tab:ablation}
\label{tab:action_conditioning}
\small
\begin{minipage}[t]{0.54\linewidth}
\vspace{0pt}
\centering
\textbf{\strut (a) Policy performance}\par\smallskip
\setlength{\tabcolsep}{3pt}
\begin{tabular}{@{}llr@{}}
\toprule
Benchmark & Configuration & Score $\uparrow$ \\
\midrule
\multirow{4}{*}{LIBERO}
& StarVLA-OFT & 96.6 \\
& + Semantic 3D prediction & 97.3 \\
& + Physical-consistency closure & 97.9 \\
& + Action conditioning & \textbf{98.3} \\
\midrule
\multirow{5}{*}{CALVIN}
& Full & \textbf{3.73} \\
& w/o kinematics & 3.68 \\
& w/o background & 3.65 \\
& w/o rigidity & 3.55 \\
& w/o closure & 3.45 \\
\bottomrule
\end{tabular}
\end{minipage}\hfill
\begin{minipage}[t]{0.435\linewidth}
\vspace{0pt}
\centering
\textbf{\strut (b) Future geometry}\par\smallskip
\setlength{\tabcolsep}{2.5pt}
\begin{tabular}{@{}llrr@{}}
\toprule
& Metric & Full & w/o action \\
\midrule
\multirow{3}{*}[-5pt]{LIBERO}
& \rule[-5.5pt]{0pt}{15pt}Inst-CD (m) $\downarrow$ & \textbf{0.0088} & 0.0124 \\
& \rule[-5.5pt]{0pt}{15pt}Trans (cm) $\downarrow$ & \textbf{1.06} & 2.59 \\
& \rule[-5.5pt]{0pt}{15pt}Recall@2cm $\uparrow$ & \textbf{0.940} & 0.884 \\
\midrule
\multirow{3}{*}[-5pt]{CALVIN}
& \rule[-5.5pt]{0pt}{15pt}Inst-CD (m) $\downarrow$ & \textbf{0.0137} & 0.0183 \\
& \rule[-5.5pt]{0pt}{15pt}Trans (cm) $\downarrow$ & \textbf{1.25} & 1.77 \\
& \rule[-5.5pt]{0pt}{15pt}Recall@2cm $\uparrow$ & \textbf{0.928} & 0.799 \\
\bottomrule
\end{tabular}
\end{minipage}
\end{table}

%% file: sec/5_conclusion.tex
\section{Conclusion}

We presented ForeAct3D, a framework for future world modeling in semantic 3D space that combines structured scene prediction, physical-consistency closure, and action-conditioned forecasting within VLA policies. ForeAct3D achieves 98.3\% average success on LIBERO and an average task length of 3.73 on CALVIN, with ablations supporting complementary contributions from its three components. Real-world cumulative ablations increase average success from 6.7\% to 37.8\% across three tabletop tasks, with action conditioning improving target-region retention in the evaluated trials. These results support physically constrained future-state prediction as a useful form of supervision for robot policy learning. Extending the state constraints beyond rigid scenes and using predicted futures to refine actions at inference time remain directions for future work.

%% file: sec/6_AI_Disclosure.tex
\section*{AI Use Statement}

In this work, we used generative AI tools for drafting and revising portions of the manuscript and assisting with literature discovery. We have not used generative AI tools for conducting experiments or generating synthetic datasets, and proving mathematical claims is not applicable to this work. Additionally, we used generative AI tools for language polishing, grammar correction, and improving clarity and organization. We have reviewed all AI-assisted work. Specifically, we checked technical descriptions against the implementation, verified numerical claims against experimental records, and checked references against their original sources. We take responsibility for the final content of this work, including text, claims or artifacts produced with the aid of generative AI.

%% file: sec/7_Ethics_Statement.tex
\section*{Ethics Statement}

This work studies robot manipulation in simulation and on tabletop tasks, with the aim of improving how policies represent and anticipate physical interactions. Learned policies can still produce inaccurate predictions or unsafe actions, especially outside the evaluated settings. The reported results do not establish safe operation around people or in safety-critical applications. Deployment should therefore include task-specific risk assessment, appropriate human oversight, and independent safety mechanisms. The assumptions and limitations discussed in Appendix~\ref{app:limitations} should guide the interpretation of results and any extension to new objects, environments, or robot platforms.

%% file: sec/8_Reproducibility_Statement.tex
\section*{Reproducibility Statement}

Section~\ref{sec:method} describes the scene representation, geometric prediction modules, physical-consistency constraints, and action conditioning. Section~\ref{sec:experiments} reports the manipulation experiments and ablations, including real-world demonstration counts, evaluation trials, and success criteria. Appendix~\ref{sec:benchmarks} specifies benchmark protocols, optimization settings, training schedules, batch sizes, and hardware. Appendices~\ref{app:closure_details} and~\ref{app:supervision_details} specify geometric reconstruction, rigid alignment, physical residuals, dense supervision, and validity handling. Appendix~\ref{app:evaluation} defines geometric metrics, point sampling, instance filtering, and aggregation rules, including the prediction horizons used for comparison. Appendices~\ref{app:semantic_results} and~\ref{app:action_ablation} provide detailed geometry, segmentation, localization, and action-conditioning results, with qualifications for interpreting and reproducing the evaluations.

%% file: sec/Appendix.tex
\appendix
\raggedbottom
\section{Appendix}

\subsection{Physical-Consistency Formulation}
\label{app:closure_details}

\noindent\textbf{World-coordinate reconstruction.}
To make the geometric constraints explicit, let $u=(u_x,u_y)$ denote a pixel and $\tilde u=(u_x,u_y,1)^\top$ its homogeneous coordinate. For prediction time $\tau\in\{t,t+T\}$, the agent-view depth $\hat D_\tau(u)$ defines a world point through labeled camera calibration:
\begin{equation}
\hat p_\tau(u)
=R_{\mathrm{wc}}\!\left(\hat D_\tau(u)K^{-1}\tilde u\right)
+t_{\mathrm{wc}}.
\label{eq:app_backprojection}
\end{equation}
Here, $K$ is the intrinsic matrix and $(R_{\mathrm{wc}},t_{\mathrm{wc}})$ maps camera coordinates to world coordinates. The fixed agent camera uses the same calibration at both endpoints. Consequently, background and rigidity residuals compare predicted scene geometry in one common frame; the predicted wrist-camera pose is supervised separately against its labeled pose at each endpoint.

\noindent\textbf{Residual distance.}
To specify the distance used by both geometric constraints, consider paired point sets $P=\{p_i\}_{i=1}^{N}$ and $Q=\{q_i\}_{i=1}^{N}$. We sum coordinatewise Smooth-L1 penalties and then average over points, using the transition threshold $\beta=0.05$\,m:
\begin{equation}
\begin{aligned}
d(P,Q)&=\frac{1}{N}\sum_{i=1}^{N}\sum_{j=1}^{3}h_\beta(p_{ij}-q_{ij}),\\
h_\beta(r)&=
\begin{cases}
r^2/(2\beta),& |r|<\beta,\\
|r|-\beta/2,& |r|\geq\beta.
\end{cases}
\end{aligned}
\label{eq:app_residual}
\end{equation}
This definition averages over paired points without dividing the coordinate sum by three. It therefore fixes the reduction convention as well as the quadratic-to-linear transition, which is needed to interpret the scale of the background and rigid-alignment losses consistently.

\noindent\textbf{Background correspondence.}
To identify comparable static regions, let $V_\tau$ be the labeled depth-validity mask and $S_\tau$ the ground-truth semantic map. The shared background set contains pixels valid and labeled as background at both endpoints. For sample $b$, write $P_b^M$ and $Q_b^M$ for reconstructed current and future points selected at the same pixel coordinates in a set $M$. The background objective is
\begin{equation}
\begin{aligned}
M_b^{\mathrm{bg}}&=\{u:V_t(u)\land V_{t+T}(u),\ S_t(u)=S_{t+T}(u)=0\},\\
\mathcal L_{\mathrm{bg}}
&=\frac{1}{|\mathcal B_{\mathrm{bg}}|}\sum_{b\in\mathcal B_{\mathrm{bg}}}
d\!\left(P_b^{M_b^{\mathrm{bg}}},Q_b^{M_b^{\mathrm{bg}}}\right).
\end{aligned}
\label{eq:app_background}
\end{equation}
The eligible set $\mathcal B_{\mathrm{bg}}$ contains samples with valid future targets and at least 32 shared background pixels. A pixel occupied by the robot or an object at either endpoint is excluded. All eligible samples have equal weight after their pointwise residuals are averaged, preventing samples with larger visible backgrounds from dominating the batch loss solely because of their pixel counts.

\noindent\textbf{Rigid alignment.}
To construct an instance pair, use ground-truth instance IDs $I_t$ and $I_{t+T}$ to retain valid pixels satisfying $I_t(u)=I_{t+T}(u)=k$ for object $k$. Instances with fewer than 32 shared pixels are excluded. For one retained pair, omit the sample and instance subscripts and denote the resulting points by $P=\{p_i\}$ and $Q=\{q_i\}$. Their unweighted centroids and cross-covariance are
\begin{equation}
\bar p=\frac{1}{N}\sum_i p_i,\qquad
\bar q=\frac{1}{N}\sum_i q_i,\qquad
H=\frac{1}{N}\sum_i(p_i-\bar p)(q_i-\bar q)^\top.
\end{equation}
Using the singular value decomposition $H=U\Sigma V^\top$, Kabsch alignment produces a rotation and translation with no scale parameter. The determinant correction excludes reflections, yielding a transform that minimizes the squared distance between the paired point sets:
\begin{equation}
R=V\operatorname{diag}\!\left(1,1,\det(VU^\top)\right)U^\top,
\qquad t=\bar q-R\bar p.
\label{eq:app_kabsch}
\end{equation}
We then evaluate $d(RP+t,Q)$ using Equation~\ref{eq:app_residual}. The transform is thus fitted with a squared-distance criterion and scored with Smooth-L1 residuals. Pairing follows shared pixel coordinates rather than tracked surface points; identical instance membership alone does not establish material-point correspondence when objects move between the two prediction endpoints.

\noindent\textbf{Instance and sample aggregation.}
To combine object residuals, let $\mathcal K_b$ contain the retained instances in sample $b$, and let $\mathcal B_{\mathrm{rigid}}$ contain samples with valid future targets and at least one retained instance. We average instance residuals within each sample and then average eligible samples:
\begin{equation}
\mathcal L_{\mathrm{rigid}}
=\frac{1}{|\mathcal B_{\mathrm{rigid}}|}
\sum_{b\in\mathcal B_{\mathrm{rigid}}}
\frac{1}{|\mathcal K_b|}\sum_{k\in\mathcal K_b}
w_{bk}\,d(R_{bk}P_{bk}+t_{bk},Q_{bk}).
\label{eq:app_rigid_batch}
\end{equation}
The contact weight is 0.1 when the gripper-closure signal is active and the current instance centroid is within 0.10\,m of the end-effector position, and 1 otherwise. The implementation uses the translation of the labeled wrist-camera pose as the position proxy for this test. The denominator remains the number of retained instances, not the sum of contact weights, so downweighting decreases the contribution of contact-associated residuals without renormalizing the remaining objects.

\noindent\textbf{Wrist-pose consistency.}
To compare predicted and reference wrist poses, the pose head outputs a translation $\hat x_\tau$ and two three-dimensional rotation vectors. Normalization and orthogonalization of the vectors, followed by their cross product, form $\hat R_\tau$. For reference pose $(R_\tau,x_\tau)$, the per-sample pose error is
\begin{equation}
\ell_{\mathrm{pose}}^\tau
=\frac{1}{3}\|\hat x_\tau-x_\tau\|_2^2
+\arccos\!\left(\frac{\operatorname{tr}(\hat R_\tau R_\tau^\top)-1}{2}\right).
\label{eq:app_pose}
\end{equation}
The inverse-cosine argument is clamped to $[-1+10^{-6},1-10^{-6}]$. Current and future pose errors are each averaged over their valid samples and then summed, with future validity handled as described below. Translation MSE and rotation angle have unit weights. The scene-supervision and closure terms use the same pose error; their coefficients add when supervising each predicted wrist pose.

\subsection{Dense Supervision and Validity Handling}
\label{app:supervision_details}

\noindent\textbf{Depth supervision.}
For one view and prediction time, let $e=\hat D-D$ denote the depth error and let $M$ select valid ground-truth depths satisfying $0<D<d_{\max}$. We set $d_{\max}=2.0$\,m for the agent camera and $1.0$\,m for the wrist camera; these training cutoffs are separate from the evaluation protocol in Appendix~\ref{app:evaluation}. Write $\langle x\rangle_M$ for the mean over valid pixels across the batch. The depth objective combines pointwise error and adjacent-pixel differences:
\begin{equation}
\mathcal L_D
=\langle |e|\rangle_M
+0.5\!\left(
\langle |\Delta_x e|\rangle_{M_x}
+\langle |\Delta_y e|\rangle_{M_y}\right).
\end{equation}
Here, $\Delta_x$ and $\Delta_y$ are adjacent-pixel differences, and $M_x$ and $M_y$ retain pairs whose two endpoints are valid. Each directional term uses its own valid-pair denominator, so pairs crossing invalid regions are excluded from both the error sum and the corresponding normalization.

\noindent\textbf{Semantic supervision.}
The segmentation objective combines pixelwise cross-entropy and soft Dice loss over the three classes: background, robot, and object. Let $p_{ic}$ be the softmax probability and $y_{ic}$ the one-hot target for class $c$ at valid pixel $i$. Dice statistics pool valid pixels across the batch, computing one overlap score per semantic class before averaging the three class losses:
\begin{equation}
\mathcal L_{\mathrm{Dice}}
=1-\frac{1}{3}\sum_{c=0}^{2}
\frac{2\sum_{i\in M_S}p_{ic}y_{ic}+1}
{\sum_{i\in M_S}p_{ic}+\sum_{i\in M_S}y_{ic}+1}.
\end{equation}
The semantic validity mask $M_S$ excludes invalid labels without imposing the depth cutoff. Cross-entropy is averaged over the same valid pixels, and the additive smoothing constant makes the Dice ratio well defined for empty support. Cross-entropy and Dice have unit weights, with uniform class weights; instance IDs supervise physical constraints without adding segmentation classes.

\noindent\textbf{Missing future targets and empty support.}
A sample-level validity flag gates all future targets. For depth and segmentation, it is incorporated into pixel masks before reduction, excluding invalid samples from both numerators and denominators. Future pose losses similarly average only over samples with valid future targets. An entirely invalid future batch contributes zero future supervision, while current-state and action supervision remain active. Background and rigidity return zero when no sample has sufficient shared support, using separate denominators for eligible samples.

\noindent\textbf{Training and evaluation support.}
The 32-pixel threshold above governs physical regularization, while the evaluation rules in Appendix~\ref{app:evaluation} select objects with at least 30 mask pixels and at least three valid reconstructed points. Training depth cutoffs are not applied to the reported geometric metrics. Geometric evaluation first averages inference windows within an episode and then weights episodes equally, whereas dense supervision pools valid pixels across the training batch. These reductions preserve the distinction between pixel-level supervision and episode-level evaluation.

\subsection{Benchmarks and Implementation Details}
\label{sec:benchmarks}

\noindent\textbf{LIBERO.}
We evaluate on four LIBERO task suites: Spatial, Object, Goal, and Long. Each suite contains 10 tasks, with 50 evaluation episodes per task and 500 episodes per suite. A single policy is trained across all four suites. We report per-suite success rates and their equally weighted average, covering spatial reasoning, object manipulation, goal-conditioned behavior, and long-horizon tasks.

\noindent\textbf{CALVIN.}
We adopt the ABC$\rightarrow$D setting, training on environments A, B, and C and evaluating zero-shot on the unseen environment D. Evaluation consists of 1,000 rollouts, each containing a sequence of five language-conditioned subtasks. Performance is measured by the average number of successfully completed consecutive subtasks per rollout, with a maximum value of 5. Table~\ref{tab:calvin} additionally reports success rates for completing successive subtasks in the sequence.

\noindent\textbf{Implementation details.}
We use the same training configuration for LIBERO and CALVIN, including all ablation variants. Models are trained using AdamW~\citep{loshchilov2017decoupled} with $\beta_1=0.9$, $\beta_2=0.95$, optimizer $\epsilon=10^{-8}$, and weight decay $10^{-8}$. The peak learning rate is $2.5\times10^{-5}$ for the vision-language backbone and $1.0\times10^{-4}$ for both the action module and geometric prediction heads. We apply linear warmup followed by cosine decay, with the scheduler's minimum learning-rate parameter set to $2\times10^{-7}$, and clip the global gradient norm to 1.0.

Training runs for 30,000 optimization steps, including 5,000 learning-rate warmup steps, on 16 NVIDIA A100 GPUs with 40\,GB of memory each. The batch size is 8 per GPU with no gradient accumulation, giving a global batch size of 128. Ablation variants retain these settings while omitting the specified modules and their associated losses; absent prediction heads receive no supervision. Physical constraints and dense supervision are defined in Appendices~\ref{app:closure_details} and~\ref{app:supervision_details}.

The current- and future-head weights and the overall scene-supervision multiplier are all 1.0. The three closure terms have unit relative weights. Their overall coefficient increases linearly from 0 to 0.1 over the first 2,000 training iterations and remains at 0.1 thereafter; this schedule is separate from the learning-rate warmup for the backbone and prediction heads.

\subsection{Geometric Evaluation Protocol}
\label{app:evaluation}

\noindent\textbf{Views and prediction horizons.}
All geometric metrics are evaluated from the agent camera and compared with ground truth at the corresponding timestep. Table~\ref{tab:geometry_full} reports current and future states separately; only future predictions are evaluated for CLOVER and DreamVLA. ForeAct3D uses a prediction horizon of $T=8$, CLOVER uses its original checkpoint with $T=5$, and DreamVLA uses $H=3$. These comparisons characterize prediction accuracy at each model's evaluation horizon and do not isolate performance differences under a shared horizon.

\noindent\textbf{Depth evaluation.}
Depth metrics use pixels where both ground-truth and predicted depth exceed $10^{-3}$\,m. All valid regions, including background, contribute without near/far clipping or foreground filtering. Ground-truth depth is resized by nearest-neighbor interpolation to the prediction resolution. Predictions are evaluated in meters without median scaling or scale-and-shift alignment. We report absolute relative error (AbsRel), root mean squared error (RMSE), and depth accuracy $\delta_1$.

\noindent\textbf{Point sampling and Chamfer distance.}
Scene Chamfer distance (Scene CD) uses 2,048 points sampled uniformly at random from each point cloud. Instance Chamfer distance (Inst-CD) uses all valid points within each ground-truth object mask without downsampling. For point sets $A$ and $B$, we compute the symmetric mean of unsquared Euclidean nearest-neighbor distances:
\begin{equation}
\mathrm{CD}(A,B)
=
\frac{1}{2}
\left(
\frac{1}{|A|}
\sum_{a\in A}
\min_{b\in B}\|a-b\|_2
+
\frac{1}{|B|}
\sum_{b\in B}
\min_{a\in A}\|b-a\|_2
\right).
\end{equation}
Both metrics are reported in meters: Scene CD measures agreement over the sampled scene reconstruction, while Inst-CD measures geometry within ground-truth object regions.

\noindent\textbf{Earth Mover's Distance.}
Scene Earth Mover's Distance (Scene EMD) uses 1,024 randomly sampled points from each cloud with a fixed random seed of 0. We compute EMD by solving the discrete optimal transport problem between the predicted and ground-truth point clouds. The cost matrix contains unsquared Euclidean distances, and each cloud has uniform point weights summing to one. Mean transport distance is reported in meters without normalizing coordinates or costs.

\noindent\textbf{Instance selection and localization metrics.}
Ground-truth instance masks select the same valid pixel locations from predicted and ground-truth depth, defining a pair of point clouds for each object. Instance association therefore uses ground-truth IDs rather than predicted instance labels or geometric matching. These metrics assess depth-derived geometry within known object regions; predicted semantic segmentation is evaluated separately as a distinct prediction task.

We exclude object ID 0, which represents background and robot regions, and objects with fewer than 30 mask pixels. Invalid predicted-depth pixels are removed from both point clouds. Instances with fewer than three remaining valid points are excluded from the metric averages and recall denominator without penalty. Translation error is the Euclidean distance between the predicted and ground-truth centroids computed from the retained points. Within each evaluation window, Recall@2cm is the fraction of retained instances with centroid error strictly below 2\,cm. It therefore measures localization among evaluable instances rather than coverage of all ground-truth objects.

\noindent\textbf{Metric aggregation.}
Metrics are first averaged over inference windows within each episode, with approximately one window per action chunk. A window that is invalid for a particular metric is excluded only from that metric's average. Episode means are then averaged with equal weight, regardless of the number of valid windows. Recall@2cm is consequently an average of window-level fractions rather than a pooled ratio over all retained instances. LIBERO summary results use an equally weighted average across all four benchmark task suites.

\begin{table}[t]
\centering\footnotesize
\caption{\textbf{Complete depth-derived geometry on CALVIN.} Instance metrics use GT masks. Rows separate current and future states; future horizons are 5 (CLOVER), 3 (DreamVLA), and 8 (ForeAct3D). Depth RMSE, CD, and EMD use meters; translation uses centimeters; recall is a fraction.}
\label{tab:geometry_full}
\setlength{\tabcolsep}{3pt}
\begin{tabular}{@{}llrrrrrrrr@{}}
\toprule
& & \multicolumn{3}{c}{Depth} & \multicolumn{2}{c}{Scene} & \multicolumn{3}{c}{GT instance regions} \\
\cmidrule(lr){3-5}\cmidrule(lr){6-7}\cmidrule(lr){8-10}
Method & State & AbsRel & RMSE & $\delta_1$ & CD & EMD & Inst-CD & Trans & Recall \\
\midrule
CLOVER & Future & 0.0083 & 0.1252 & 0.9955 & 0.0195 & 0.0601 & 0.0139 & 1.741 & 0.7582 \\
DreamVLA & Future & 0.0867 & 0.5583 & 0.9126 & 0.1350 & 0.3145 & 0.1455 & 20.61 & 0.1073 \\
ForeAct3D & Current & 0.0110 & 0.1551 & 0.9942 & 0.0206 & 0.0631 & 0.0129 & 1.165 & 0.9416 \\
ForeAct3D & Future & 0.0117 & 0.1633 & 0.9933 & 0.0206 & 0.0623 & 0.01371 & 1.245 & 0.9278 \\
\bottomrule
\end{tabular}
\end{table}
\begin{table}[t]
\centering\small
\caption{\textbf{Three-class semantic segmentation quality (\%).} Future class IoUs are reported for background, robot, and object. All objects share the same semantic class. LIBERO Average denotes the mean across the Object, Goal, Spatial, and Long task suites.}
\label{tab:semantic_quality}
\begin{tabular}{@{}llrrrrr@{}}
\toprule
Benchmark & Suite & mIoU@$t$ & BG@$t+T$ & Robot@$t+T$ & Object@$t+T$ & mIoU@$t+T$ \\
\midrule
LIBERO & Object & 98.0 & 99.4 & 95.8 & 98.0 & 97.7 \\
& Goal & 87.2 & 94.8 & 94.8 & 70.8 & 86.8 \\
& Spatial & 78.1 & 91.5 & 95.7 & 45.7 & 77.6 \\
& Long & 93.7 & 97.2 & 94.5 & 87.8 & 93.2 \\
& Average & 89.3 & 95.7 & 95.2 & 75.6 & 88.8 \\
\midrule
CALVIN & ABC$\rightarrow$D & 88.0 & 89.8 & 90.3 & 77.3 & 85.8 \\
\bottomrule
\end{tabular}
\end{table}

\subsection{Detailed Geometry and Semantic Prediction Results}
\label{app:semantic_results}

\noindent\textbf{Geometry and semantic segmentation.}
Table~\ref{tab:geometry_full} reports the complete depth and reconstructed-geometry results on CALVIN. Table~\ref{tab:semantic_quality} reports semantic segmentation quality for current and future states, including future IoU for background, robot, and object regions. All objects share one semantic class. These evaluations separately characterize depth-derived geometry within ground-truth instance regions and the accuracy of the predicted semantic masks.

\noindent\textbf{Future localization with predicted masks.}
Table~\ref{tab:semantic_localization} evaluates localization at $t+T$ using predicted depth with either ground-truth or predicted semantic masks. Both settings use ground-truth instance IDs to identify objects. Predicted-mask results thus assess localization using predicted semantic support, rather than instance detection without ground-truth assistance.

The filtering and aggregation rules in Appendix~\ref{app:evaluation} describe the GT-mask geometry evaluation. Identical instance filtering across the two mask settings has not been established, and mask-dependent point selection can change both centroid estimates and the evaluated instance population. Differences between the two settings therefore cannot be attributed solely to segmentation accuracy. In particular, lower localization error with predicted masks does not by itself indicate better object coverage. LIBERO GT-mask results follow the action-conditioning protocol in Table~\ref{tab:libero_action_details}.

\begin{table}[!htbp]
\centering\small
\caption{\textbf{Future localization with predicted depth and semantic masks.} GT instance IDs are retained in both mask settings. Recall@2cm is in percent. These results assess localization with predicted semantic support, not autonomous instance detection without GT IDs.}
\label{tab:semantic_localization}
\begin{tabular}{@{}llrrrr@{}}
\toprule
& & \multicolumn{2}{c}{Trans (cm) $\downarrow$} & \multicolumn{2}{c}{Recall@2cm $\uparrow$} \\
\cmidrule(lr){3-4}\cmidrule(lr){5-6}
Benchmark & Suite & GT mask & Pred mask & GT mask & Pred mask \\
\midrule
LIBERO & Object & 0.56 & 0.48 & 96.7 & 96.9 \\
& Goal & 1.58 & 1.20 & 91.2 & 92.8 \\
& Spatial & 0.71 & 0.67 & 95.7 & 96.3 \\
& Long & 1.38 & 1.13 & 92.4 & 93.8 \\
& Average & 1.06 & 0.87 & 94.0 & 94.9 \\
\midrule
CALVIN & ABC$\rightarrow$D & 1.25 & 3.49 & 92.8 & 65.9 \\
\bottomrule
\end{tabular}
\end{table}

\subsection{Complete Action-Conditioning Results}
\label{app:action_ablation}

\noindent\textbf{CALVIN.}
Table~\ref{tab:action_conditioning_full} reports future geometry with and without action conditioning. Both variants use the same backbone, geometric queries, physical-consistency closure, training data, evaluation protocol, and prediction horizon $T=8$. They differ only in whether the future branch receives the policy-generated action chunk. Action conditioning improves all reported metrics, with larger relative reductions in object-level errors than in scene-level errors.

\begin{table}[!htbp]
\centering
\caption{Effect of action conditioning on future depth-derived geometry on CALVIN. Both variants use the same evaluation protocol and prediction horizon of $T=8$.}
\label{tab:action_conditioning_full}
\begin{tabular}{lcc}
\toprule
Metric & ForeAct3D & w/o Action Conditioning \\
\midrule
AbsRel $\downarrow$ & 0.0117 & 0.0165 \\
RMSE (m) $\downarrow$ & 0.1633 & 0.2144 \\
$\delta_1$ $\uparrow$ & 0.9933 & 0.9881 \\
Scene CD (m) $\downarrow$ & 0.0206 & 0.0220 \\
Scene EMD (m) $\downarrow$ & 0.0623 & 0.0655 \\
Inst-CD (m) $\downarrow$ & 0.0137 & 0.0183 \\
Trans (cm) $\downarrow$ & 1.25 & 1.77 \\
Recall@2cm $\uparrow$ & 0.928 & 0.799 \\
\bottomrule
\end{tabular}
\end{table}

\noindent\textbf{LIBERO.}
Table~\ref{tab:libero_action_details} reports depth and instance-level metrics for the Goal, Object, Spatial, and Long suites, with the Long suite denoted as LIBERO-10 in the detailed results. Action conditioning improves every reported metric across all four suites. The LIBERO values in Table~\ref{tab:ablation}(b) are equally weighted averages of the suite-level results, computed before rounding. These per-suite results complement the aggregate comparison across manipulation settings.

\begin{table}[htbp]
\centering
\small
\caption{\textbf{Action-conditioning results by LIBERO suite.} Values are measured at the future prediction horizon under the same GT-mask protocol as Table~\ref{tab:semantic_localization}. Full denotes ForeAct3D with action conditioning; w/o action denotes the variant without it. The main-text instance metrics are equal-weight averages of the four suites, computed before rounding.}
\label{tab:libero_action_details}
\setlength{\tabcolsep}{3pt}
\begin{tabular}{@{}llrrrrrr@{}}
\toprule
Suite & Variant & AbsRel $\downarrow$ & RMSE $\downarrow$ & $\delta_1$ $\uparrow$ & Inst-CD $\downarrow$ & Trans $\downarrow$ & Recall $\uparrow$ \\
& & & (m) & & (m) & (cm) & @2cm \\
\midrule
\multirow{2}{*}{Goal} & Full & 0.0225 & 0.214 & 0.976 & 0.0101 & 1.58 & 0.912 \\
 & w/o action & 0.0562 & 0.315 & 0.945 & 0.0126 & 2.65 & 0.879 \\
\midrule
\multirow{2}{*}{Object} & Full & 0.0078 & 0.067 & 0.994 & 0.0068 & 0.56 & 0.967 \\
 & w/o action & 0.0388 & 0.216 & 0.958 & 0.0115 & 2.82 & 0.908 \\
\midrule
\multirow{2}{*}{Spatial} & Full & 0.0229 & 0.230 & 0.973 & 0.0076 & 0.71 & 0.957 \\
 & w/o action & 0.0769 & 0.378 & 0.929 & 0.0123 & 2.33 & 0.880 \\
\midrule
\multirow{2}{*}{LIBERO-10} & Full & 0.0173 & 0.095 & 0.987 & 0.0105 & 1.38 & 0.924 \\
 & w/o action & 0.0598 & 0.224 & 0.951 & 0.0133 & 2.55 & 0.869 \\
\bottomrule
\end{tabular}
\end{table}

\subsection{Limitations and Future Directions}
\label{app:limitations}

ForeAct3D assumes static backgrounds and rigid instances, limiting applicability to deformable objects, articulated instances, and moving backgrounds~\citep{wang2026raymap3r,wang2026cognimap3d}. Same-pixel instance membership approximates temporal correspondence without tracking material points and becomes less reliable under large motion. Training requires dense geometric supervision and instance annotations, making predictions sensitive to depth and mask errors. Broader task and robot coverage are needed to assess generalization and variability. Geometric comparisons use different prediction horizons, and predicted-mask localization retains ground-truth instance IDs. Future prediction provides training supervision but is unused at inference. Future work could reduce supervision, extend physical constraints to general scenes, and use predicted futures to evaluate or refine actions.